\documentclass[
]{ceurart}

\usepackage{listings}

\usepackage{graphicx}
\usepackage{amsmath}
\usepackage{amssymb}
\usepackage{booktabs}

\usepackage{subcaption}
\usepackage{caption}

\usepackage{latexsym}
\usepackage{amsthm}
\usepackage{enumitem}
\usepackage{color}
\usepackage{amsopn, amscd}
\usepackage{mathtools}
\usepackage{amsfonts} 
\usepackage[ruled]{algorithm2e}

\providecommand{\dd}{\mathrm{d}}
\providecommand{\exp}{\mathrm{exp}}

\providecommand{\R}{\mathbb{R}}

\providecommand{\C}{\mathbb{C}}

\providecommand{\sl}{\mathfrak{sl}}
\providecommand{\Hess}{\mathrm{Hess}}

\newcommand{\argmax}{\mathop{\mathrm{argmax}}}
\newcommand{\argmin}{\mathop{\mathrm{argmin}}}

\newtheorem{theorem}{Theorem}
\newtheorem{lemma}[theorem]{Lemma}
\newtheorem{corollary}[theorem]{Corollary}

\newtheorem{definition}{Definition}

\begin{document}

\copyrightyear{2024}
\copyrightclause{Copyright for this paper by its authors.
  Use permitted under Creative Commons License Attribution 4.0
  International (CC BY 4.0).}

\conference{32nd Irish Conference on Artificial Intelligence and Cognitive Science, December 09--10, 2024, Dublin, Ireland}

\title{Tangentially Aligned Integrated Gradients for User-Friendly Explanations}

\author[1]{Lachlan Simpson}[%
email = lachlan.simpson@adelaide.edu.au]
\cormark[1]
\author[2]{Federico Costanza}
\author[3]{Kyle Millar}
\author[1,3]{Adriel Cheng}
\author[1]{Cheng-Chew Lim}
\author[1]{Hong Gunn Chew}

\address[1]{School of Electrical and Mechanical Engineering, The University of Adelaide, Australia}

\address[2]{School of Computer and Mathematical Sciences, The University of Adelaide, Australia}

\address[3]{Information Sciences Division, Defence Science and Technology Group, Australia}

\cortext[1]{Corresponding author.}

\begin{abstract}
Integrated gradients is prevalent within machine learning to address the black-box problem of neural networks. The explanations given by integrated gradients depend on a choice of base-point. The choice of base-point is not a priori obvious and can lead to drastically different explanations. There is a longstanding hypothesis that data lies on a low dimensional Riemannian manifold. The quality of explanations on a manifold can be measured by the extent to which an explanation for a point lies in its tangent space. In this work, we propose that the base-point should be chosen such that it maximises the tangential alignment of the explanation. We formalise the notion of tangential alignment and provide theoretical conditions under which a base-point choice will provide explanations lying in the tangent space. We demonstrate how to approximate the optimal base-point on several well-known image classification datasets. Furthermore, we compare the optimal base-point choice with common base-points and three gradient explainability models.
\end{abstract}

\maketitle

\section{Introduction}
\label{sec:intro}

Deep learning provides state-of-the-art solutions to a wide array of computer vision tasks \cite{yolo}. The accuracy of deep learning comes with the trade-off of interpretability \cite{zednik2019solving}. A fundamental problem of deep learning is how a model reached a prediction \cite{Sejnowski_2020}. Post hoc gradient explainability models address the black-box problem by providing an attribution of the input features to the prediction of neural network under analysis \cite{simpson2024probabilistic}. Several gradient explainability methods exist with the underlying assumption that analysis of the model's gradient highlights features with greatest impact on a prediction \cite{smilkov2017smoothgrad,pmlr-v70-sundararajan17a}.

Several metrics have been proposed to measure the quality of explainability models. In \cite{simpson2024probabilistic,khan2023analyzing}, the authors propose the Lipschitz constant of an explainability model as a measure of explainability quality. Other works consider the extent to which an explainability model approximates the underlying neural network as a measure of quality. These metrics do not consider the user's perception of the explanations. Following from Ganz et al.'s \cite{ganz} notion of perceptually aligned gradients of a neural network, Brodt et al. \cite{brodt} introduce perceptually aligned explanations. Brodt el al. \cite{brodt} measure how perceptually aligned an explanation is by the extent to which an explanation lies in the tangent space of the manifold. Brodt el al. \cite{brodt}'s measure of tangential explanations relies on the manifold hypothesis. The manifold hypothesis is the notion that data lies on a low dimensional Riemannian manifold \cite{whiteley2024statisticalexplorationmanifoldhypothesis,fefferman2013testing,GoodBengCour16, ganz, tsipras2018robustness,madry2018towards}. 

The tangent space captures the features of an image that can be changed whilst remaining in the distribution of images. The intuition is if an explanation lies in the tangent space of the image, the explanation will contain meaningful components of the image \cite{brodt}. Brodt et al. \cite{brodt} demonstrate their hypothesis on several gradient explainability models on well-known computer vision datasets. Brodt el al. \cite{brodt} further demonstrate tangentially aligned explanations are robust to adversarial attacks.

Integrated gradients (IG) \cite{pmlr-v70-sundararajan17a} is a popular explainability method employed in a wide array of computer vision tasks \cite{das2020opportunities}. IG relies on a hyper-parameter known as the base-point. The choice of base-point fundamentally alters the explanation provided \cite{kindermans2017unreliability}. Base-point selection is domain dependent and chosen heuristically. The zero vector, however, is a prevalent choice in computer vision, NLP and graph machine learning \cite{pmlr-v70-sundararajan17a,pmlr-v196-xenopoulos22a,NEURIPS2020_417fbbf2}.
Several works have investigated different choices of base-point, however, none are able to determine a correct choice \cite{drakard2022exploring}. In this work we investigate the conditions under which a choice of base-point will provide perceptually aligned explanations.

The contributions of this work are twofold:
 \begin{enumerate}
     \item We provide sufficient conditions for when integrated gradient explanations are tangentially aligned. We extend these results to any base-point attribution method.
     \item 
        We provide a framework to choose a base-point point which provides meaningful explanations to the user. We compare our method with three gradient explainability models and IG with common base-points. We demonstrate that our base-point choice provides better tangential alignment and consequently more meaningful explanations. We validate our approach on four well-known computer vision datasets. 
 \end{enumerate}
The remainder of this work is structured as follows: Section 2 provides related work and background. Section 3 investigates theoretical conditions for tangential alignment of base-point attribution methods. Section 4 calculates base-points for tangential alignment of IG on four well known datasets. We compare tangential IG with four common base-point choices and three gradient explainability models. We conclude in Section 5 with a discussion for future works.

\section{Related Work and Background}

\subsection{Tangentially Aligned Integrated Gradients Explanations}
Post hoc explainability models are methods for providing an attribution for the features that influence the output of a neural network. Post hoc explainability is a step towards addressing the black-box problem
[6]. 

Base-point attribution methods (BAM) \cite{lundstrom2022rigorous} are a specific class of post hoc explainability models. A BAM is a function 
\begin{align}
 A \colon &M \times M \times \mathcal{F}(M) \to \mathbb{R}^{d}\\
 &(x,x',F) \mapsto A(x,x',F)
\end{align}
 where, $M \subset \R^d$ is a manifold, $\mathcal{F}(M)$ denotes the set of neural networks on $M$ and $x,x' \in M$ are an input and a base-point, respectively. 

We will further restrict the space of BAM functions to path methods, and we will generalise the definition of path methods to be independent of coordinates. Given a closed interval $I := [a, b] \subset \R$, a path $\gamma \colon I \to M$ and a unit vector $v \in \R^{d}$, the component of a path method $A^{\gamma} : M \times M \times \mathcal{F}(M) \to \R^{d}$ in the direction of $v$ is defined as
\begin{equation}
    A^{\gamma}_{v}(x,x',F) = \int_{a}^{b} \langle \nabla F(\gamma(t)), v \rangle \langle \gamma'(t), v \rangle \dd t.
\end{equation}
In this way, for a given orthonormal basis $\lbrace v_{1}, \dots, v_{d} \rbrace$ of $\R^{d}$, $A^{\gamma}$ is expressed as
\begin{equation}
    A^{\gamma}(x,x',F) = \sum\limits_{i = 1}^{d} A^{\gamma}_{v_{i}}(x,x',F) v_{i}.
\end{equation}
Particularly, for the standard orthonormal basis $\lbrace e_{1}, \dots, e_{d} \rbrace$ of $\R^{d}$, we obtain the usual definition
\begin{equation}
\label{eqn:gen_path}
    A^{\gamma}_{e_{i}}(x,x',F) = \int_{a}^{b} \frac{\partial F}{\partial x_{i}}(\gamma(t)) \frac{\partial \gamma_{i}}{\partial t}(t) \dd t.
\end{equation}
The prominent path method, integrated gradients \cite{pmlr-v70-sundararajan17a} is a path method where $\gamma$ is taken to be the straight line between points $x,x' \in M$. For any pair of points $x, x' \in M$, a neural network $F \in \mathcal{F}(M)$, and a unit vector $v$, integrated gradients of the $v$ component of $x$ is defined to be:
\begin{equation}
\mathrm{IG}_{v}(x, x', F) : = \langle x - x', v \rangle  \int_{0}^{1} \langle \nabla F (x' + t(x - x')), v \rangle \dd t.
\end{equation}
Letting $\mathrm{I} : M \times M \times \mathcal{F}(M) \to \R^{n}$ be the map defined by
\begin{equation}
\mathrm{I}(x, x', F) : = \int_{0}^{1} (\nabla F)(x' + t(x - x')) \dd t,
\end{equation}
integrated gradients can be expressed succinctly in the standard orthonormal basis of $\R^{d}$ as 
\begin{equation}\label{def:integrated_gradients_attribution}
\mathrm{IG}(x, x', F) = (x - x') \odot \mathrm{I}(x, x', F),
\end{equation}
where $\odot$ denotes the Hadamard product.

Several metrics have been proposed to measure the quality of explainability models. In \cite{khan2023analyzing,simpson2024probabilistic}, Lipschitzness is proposed as a measure of explainability quality. Other works consider the extent an explainability model approximates the neural network as a measure of quality. Brodt et al. \cite{brodt} propose the extent to which an explanation lies in the tangent space of the manifold as a measure of explanation quality. Attributions which lie in tangent space were demonstrated to
constitute the meaningful features that contribute to a prediction \cite{ganz,brodt}. Orthogonal attributions were closer to random noise.
The hypothesis that tangential explanations provide meaningful explanations is validated on several image classification datasets and a user study \cite{brodt}. Here tangentially aligned explanations is formalised. 

For the reminder of this work we will consider $\R^{d}$, equipped with its standard inner product $\langle \cdot, \cdot \rangle$, and we will let $M \subset \R^{d}$ be a manifold of dimension $n < d$. We will also write $\langle \cdot, \cdot \rangle$ for the restriction of the inner product of $\R^{d}$ to $M$, such that $(M, \langle \cdot, \cdot \rangle)$ is an embedded Riemannian submanifold of $(\R^{d}, \langle \cdot, \cdot \rangle)$. We will denote the tangent space of $M$ at a point $x$ by $T_{x}M$ which, in the context of data manifolds, will consist of all $v \in \R^{d}$ such that $x + v$ is ``close" to $M$, with $\|v\|_{2}$ small \cite{brodt}. Lastly, making use of the inner product of $\R^{d}$,  for each $x \in M$ we have orthogonal direct sum decomposition $T_{x}\R^{d} = T_{x}M \oplus T_{x}M^{\perp}$,where
\begin{equation}
 T_{x}M^{\perp} : = \lbrace u \in T_{x}\R^{d} \; : \; \langle u, v \rangle = 0, \; \forall v \in T_{x}M \rbrace.  
\end{equation}
We will let $\pi_{x} : T_{x}\R^{d} \to T_{x}M$ denote the natural projection from $T_{x}\R^{d}$ to $T_{x}M$ defined by
\begin{equation}\label{eqn:projections}
    \pi_{x} (v) = \sum\limits_{\ell=1}^{n} \langle v, \tau_{\ell} \rangle \tau_{\ell},
\end{equation}
where, $\lbrace \tau_{1}, \dots \tau_{n} \rbrace$ is an orthonormal basis for $T_{x}M$. We define the map $\mu_{x} : T_{x}\R^{d} \to [0, 1]$, given by
\begin{equation}\label{def:mu}
    \mu_{x} (v) : = \frac{\| \pi_{x}v \|_{2}^{2}}{\| v \|_{2}^{2}}, \quad \quad v \in T_{x}\R^{d}.
\end{equation}
The map defined in Equation (\ref{def:mu}) provides us a measure of ``how much" of a vector lies in the tangent space of $M$ at $x$, i.e. a vector $v$ is in $T_{x}M$ if and only if $\mu_{x}(v) = 1$ and, on the other hand, $v$ will be in $T_{x}M^{\perp}$ if and only if $\mu_{x}(v) = 0$, which can be observed directly from its definition. Moreover, letting $\pi_{x}^{\perp} : T_{x}\R^{d} \to T_{x}M^{\perp}$ denote the natural projection and, noting that, 
\begin{align}
   &v = \pi_{x}v + \pi_{x}^{\perp}v,\\
   &\|v\|_{2}^{2} = \| \pi_{x}v \|_{2}^{2} + \| \pi_{x}^{\perp}v \|_{2}^{2},
\end{align}
we can express $\mu_{x}$ as
\begin{equation}\label{def:mu2}
    \mu_{x} (v) : = \frac{\| \pi_{x}v \|_{2}^{2}}{\| \pi_{x}v \|_{2}^{2} + \| \pi_{x}^{\perp}v \|_{2}^{2}}, \quad \quad v \in T_{x}\R^{d}.
\end{equation}
Minimising the norm of the projection into $T_{x}M^\perp$ provides a framework to ensure tangential alignment.

\subsection{Base-point Selection for Integrated Gradients}
\label{sec:common_choices}
The attribution of IG depends on the base-point chosen. Base-point selection is domain dependent and chosen heuristically. Here we review common base-point choices as provided by \cite{sturmfels2020visualizing}.


\begin{enumerate}
    \item \textbf{Zero}. Here the base-point for all points is a constant zero vector
    \begin{equation}
        \label{eqn:zero}
        \alpha^{\text{zero}} = 0.
    \end{equation}
    in general the zero base-point can be any constant vector. 
    \item \textbf{Maximum Distance}. For a given input $x \in M$, $\alpha$ is defined as the point in $M$ of maximum distance from $x$ i.e. 
    \begin{equation}
    \label{eqn:max_dist}
        \alpha_{x}^{\max} = \argmax_{y \in M} \|x-y\|_{p}.
    \end{equation}
    Usually $p = 1$ or $2$.
    
    \item \textbf{Uniform}. We sample uniformly over a valid range of $M$
    \begin{equation}
    \label{eqn:uniform}
    \alpha^{\text{uniform}}_{i} \sim U(\min_{i},\max_{i}).
    \end{equation}

    \item \textbf{Gaussian}. A Gaussian filter is applied to the input $x$. 
    \begin{equation}
        \label{eqn:gauss}
        \alpha^{\text{Gaussian}} = \sigma \cdot v+x,
    \end{equation}
    where, $v_{i} \sim \mathcal{N}(0,1)$ and $\sigma \in \mathbb{R}$.
    We require the $\alpha^{\text{Gaussian}}$ is still within the data distribution so $\alpha^{\text{Gaussian}} \to \alpha^{\text{Uniform}}$ as $\sigma \to \infty$ \cite{sturmfels2020visualizing}. 
\end{enumerate}

The zero base-point (Equation \ref{eqn:zero}) will not highlight the aspects of the image which may be important if the object of interest contains black pixels \cite{sturmfels2020visualizing,sundararajan2018noteaboutlocalexplanation}. To address the issue of a constant base-point missing important features maximum distance (Equation \ref{eqn:max_dist}) was proposed in \cite{sturmfels2020visualizing}. Maximum distance takes the furthest point (in $\ell_{p}$ distance) from the input image such that the base-point does not contain important information of the input. Another alternative is to sample a base-point from a distribution such as uniform (Equation \ref{eqn:uniform}) or Gaussian (Equation \ref{eqn:gauss}) \cite{sturmfels2020visualizing,Fong_2017}. Despite the various choices of base-point we demonstrate none of the aforementioned base-points provide perceptually aligned explanations. 


 Zaher et al. \cite{zaher2024manifoldintegratedgradientsriemannian} propose Manifold Integrated Gradients (MIG). MIG replaces the straight line in IG with a geodesic such that the attribution lies in the Riemannian manifold. Whilst MIG addresses the problem of IG not conforming to the geometry of the data. MIG does not resolve the issue of base-point choice nor does MIG ensure that the attribution lies in the tangent space of the manifold.

 \section{Optimising the Base-point for Tangentially Aligned Explanations}

Throughout this section, we will study the map defined in Equation \ref{def:mu}, to identify possible choices of base-points for the attribution given by a BAM to be tangent to $M$ at a point. To be precise, for a given BAM,
we want to find $\alpha \in M$ such that the map
\begin{equation}\label{eqn:map1}
    x' \mapsto \mu_{x}(A(x, x', F))
\end{equation}
attains its maximum and, particularly, when this maximum value is equal to 1. We note that $\alpha = x$ is always a solution, however, we will always require $\alpha \neq x$ for non-trivial solutions.
\begin{definition}
    Let $A : M \times M \times \mathcal{F}(M) \to \R^{d}$ be a BAM and $x, \alpha \in M$, $F \in \mathcal{F}(M)$. $A$ is \textit{tangentially aligned} at $x$, with base-point $\alpha$, if $\mu_{x}(A(x, \alpha, F)) = 1$.
\end{definition}
In the remainder of this section $x \in M$ and $F \in \mathcal{F}(M)$ will be fixed, unless otherwise stated. Letting $\pi_{x}^{\perp} : T_{x}\R^{d} \to T_{x}M^{\perp}$ denote the natural projection and defining the maps
\begin{equation}\label{def:H}
H_{x}:M \to T_{x}M^{\perp}, \quad H_{x}(x') : = \pi_{x}^{\perp} A(x, x', F)
\end{equation}
and
\begin{equation}\label{def:E}
E_{x} : M \to \R, \quad E_{x}(x') := \frac{1}{2} \| H_{x}(x') \|_{2}^{2},
\end{equation}
we can characterise tangentially aligned BAM explanations with the following theorem.

\begin{theorem}\label{prop:tangentially_aligned}
    Let $A : M \times M \times \mathcal{F}(M) \to \R^{d}$ be a BAM and $x, \alpha \in M$, $F \in \mathcal{F}(M)$. Then $A$ is tangentially aligned at $x$, with base-point $\alpha$, if and only if $H_{x}(\alpha) = 0$ or, equivalently, if $E_{x}(\alpha) = 0$.
\end{theorem}
\begin{proof}
    It is immediate from the definitions of $H_{x}$ and $E_{x}$, since they are the projection to $T_{x}M^{\perp}$ of $A$ and a multiple of its norm, respectively.
\end{proof}

Choosing an orthonormal basis 
\begin{equation}
 \lbrace \tau_{1}, \dots, \tau_{n}, \nu_{n+1}, \dots, \nu_{d} \rbrace   
\end{equation}
of $T_{x}\R^{d}$ such that $\lbrace \tau_{i} \rbrace_{i = 1}^{n}$ and $\lbrace \nu_{i} \rbrace_{i = n+1}^{d}$ are orthonormal basis of $T_{x}M$ and $T_{x}M^{\perp}$, respectively, we observe that 
\begin{equation}
    \begin{array}{rcl}
    H_{x}(x') & = & A(x, x', F) - \sum\limits_{i = 1}^{n} \langle A(x, x', F), \tau_{i} \rangle \tau_{i} =  \sum\limits_{i = n+1}^{d} \langle A(x, x', F), \nu_{i} \rangle \nu_{i}\\
    \end{array}
\end{equation}
and 
\begin{equation}
    E_{x}(x') = \frac{1}{2}\sum\limits_{i = n+1}^{d} \langle A(x, x', F), \nu_{i} \rangle^{2}.
\end{equation}
Therefore, any choice of a basis for $T_{x}\R^{d}$, adapted to the splitting of $T_{x}\R^{d}$ into tangent and normal spaces of $M$ at $x$, will provide us with with a system of equations to test for tangentially aligned explanations.

Theorem \ref{prop:tangentially_aligned} provides us with a necessary condition that a base-point must satisfy to obtain a tangentially aligned explanation. To observe this, suppose that there exists $\alpha \in M$ such that $A(x, \alpha, F)$ is tangentially aligned. Then, by Theorem \ref{prop:tangentially_aligned}, $E_{x}(\alpha) = 0$ and since $E_{x}(x') \geq 0$ for all $x' \in M$, it is in fact a global minimum of $E_{x}$ and, consequently, $(\nabla E_{x})(\alpha) = 0$. Moreover, its Hessian matrix $\Hess E_{x}$ is positive definite at $\alpha$.

To simplify notation, in what follows we will denote the partial derivatives with respect to $x_{i}$ and $x_{i}'$ by $\partial_{i}$ and $\partial_{i}'$, respectively.

\begin{corollary}
    \label{cor:tang_align_necessary_condition}
    It is a necessary condition for $A(x, \alpha, F)$ to be tangentially aligned, that
    \begin{equation}
    \langle H_{x}(\alpha), (\partial_{i}'H_{x})(\alpha) \rangle = 0,
    \end{equation} 
    for all $i =1, \dots, d$.
\end{corollary}
\begin{proof}
    If $A(x, \alpha, F)$ is tangentially aligned, then $(\nabla E_{x})(\alpha) = 0$, which is equivalent to $(\partial_{i}'E_{x})(\alpha) = 0$ for all $i = 1, \dots, d$. It follows from the definition of $E_{x}$ that:
    \begin{align}
        \langle H_{x}(\alpha), (\partial_{i}'H_{x})(\alpha) \rangle &= \frac{1}{2} \partial_{i}' \langle H_{x}, H_{x} \rangle |_{\alpha} = (\partial_{i}' E_{x})(\alpha) = 0.
    \end{align}
    for all $i = 1, \dots, d$, as claimed.
\end{proof}
In order to find conditions for the Hessian matrix of $E_{x}$ to be positive definite, we will make use of \textit{Ger\v{s}gorin circle theorem} \cite{Gershgorin1931} to find bounds for the eigenvalues of $\Hess\: E_{x}$.
For a given complex $n \times n$ matrix $A$, its $i$-th \textit{Ger\v{s}gorin disk} is the closed disk $G_{i}(A) : = D(A_{ii}, R_{i}) \subset \C$, where the radius is given by the formula
\begin{equation}
R_{i} = \sum_{j \in J_{i}} | A_{ij}|, \quad J_{i} = \lbrace 1, \dots, i-1, i+1, \dots, n \rbrace.
\end{equation}


\begin{lemma}\label{lemma:Gershgorin_real_symmetric}
    Let $A$ be a real symmetric matrix such that $A_{ii} > R_{i}$ for all $i$, then $A$ is positive definite. 
\end{lemma}
Lemma \ref{lemma:Gershgorin_real_symmetric} follows immediately from \cite{Gershgorin1931}. The following theorem is an immediate consequence of Corollary \ref{cor:tang_align_necessary_condition} and of Lemma \ref{lemma:Gershgorin_real_symmetric} applied to $\Hess E_{x}$.

\begin{theorem}
    It is a sufficient condition for $A(x, \alpha, F)$ to be tangentially aligned, that for all $i$
    \begin{equation}
        \langle H_{x}(\alpha), (\partial_{i}'H_{x})(\alpha) \rangle = 0
    \end{equation}
    and that 
    \begin{equation}\label{eqn:Hessian_Gershgorin}
        (\Hess\;E_{x})(\alpha)_{ii} > R_{i}(\alpha),
    \end{equation}
    where $R_{i}(\alpha)$ denotes the radius of the $i$-th Ger\v{s}gorin disk of $(\Hess E_{x})(\alpha)$.
\end{theorem}

\begin{figure}
    \centering
    \begin{subfigure}[b]{\linewidth}
        \centering
        \includegraphics[width=\linewidth]{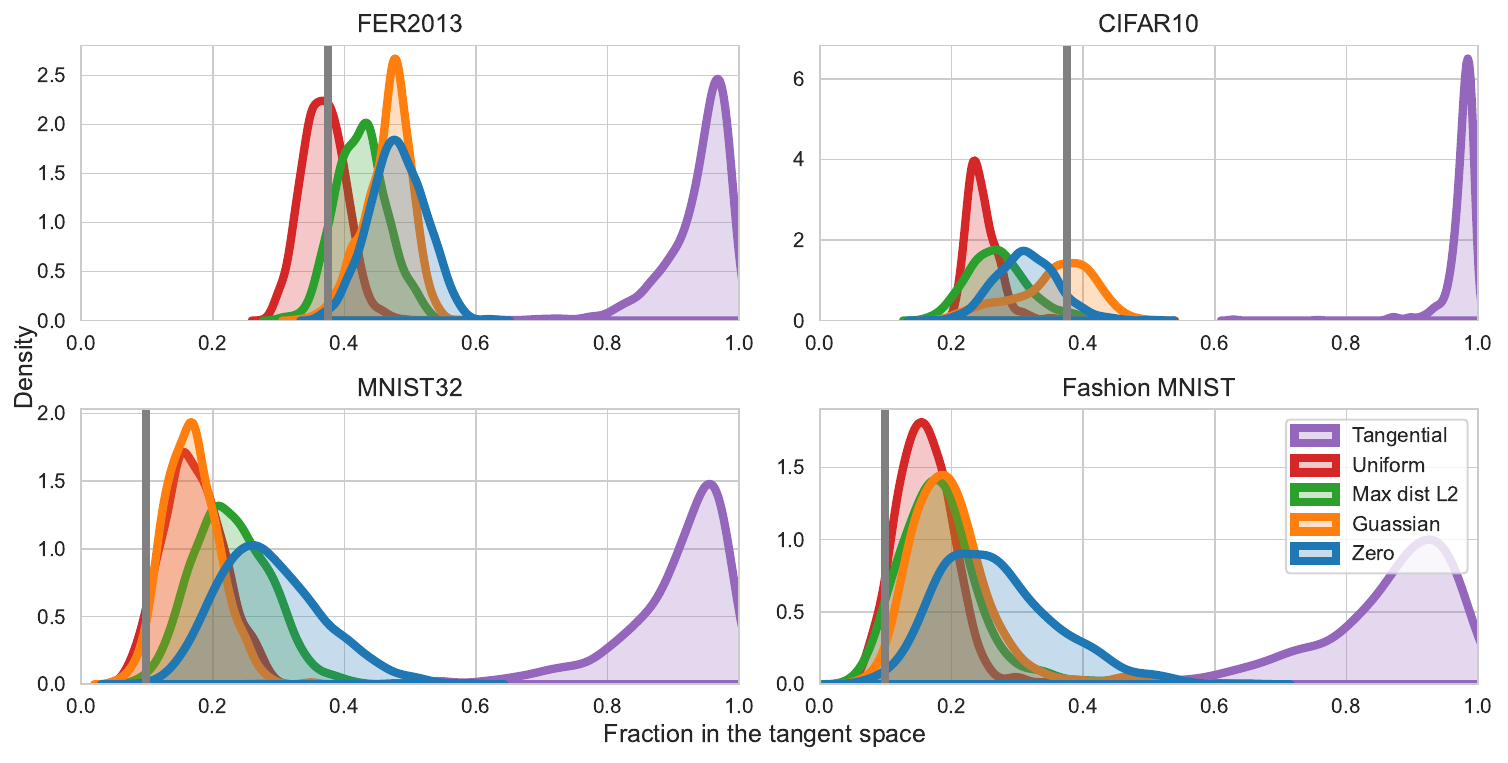}
        \caption{Base-point choices.}
        \label{fig:base-point_all_dataset}
    \end{subfigure}
    
    \vspace{10pt} 

    \begin{subfigure}[b]{\linewidth}
        \centering
        \includegraphics[width=\linewidth]{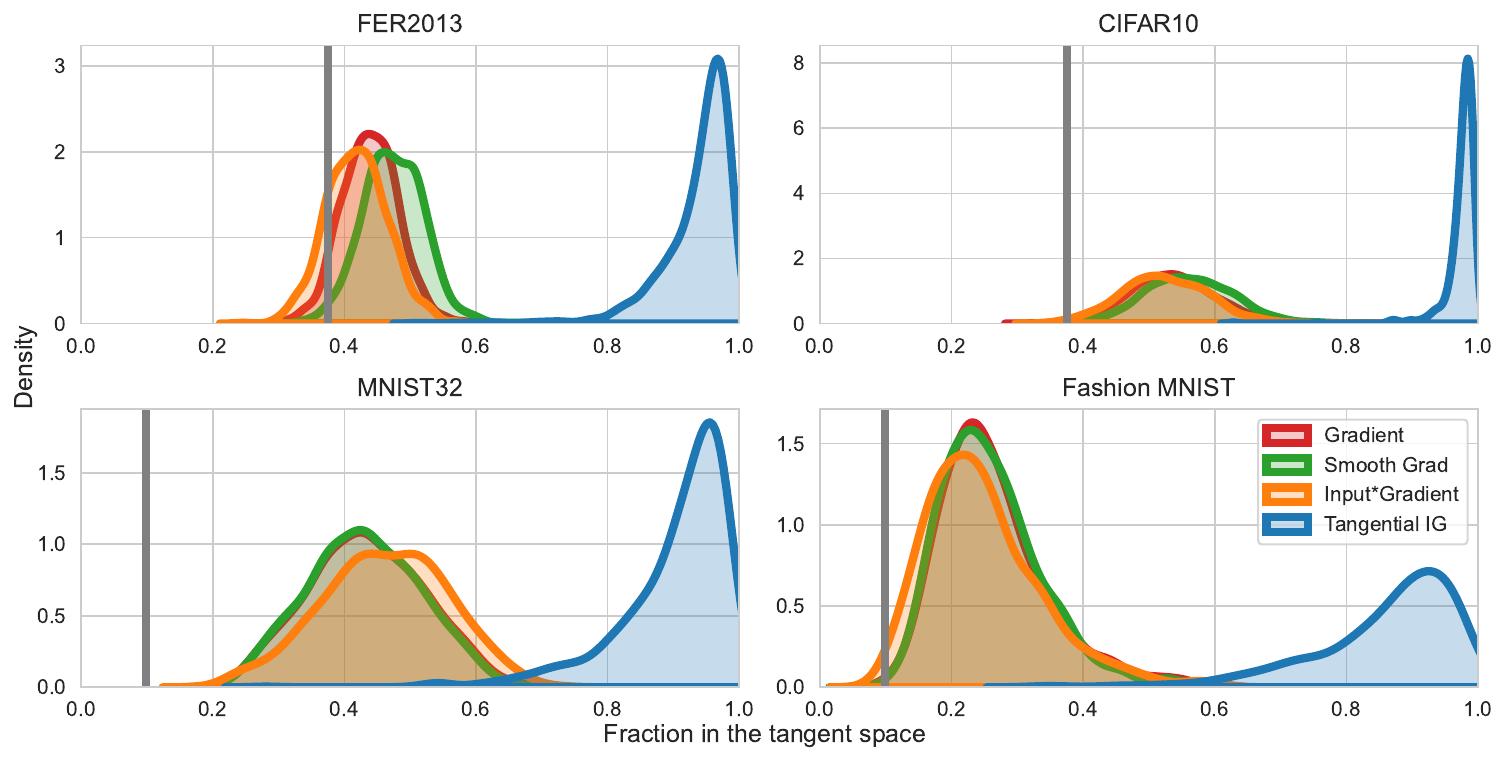}
        \caption{Gradient models.}
        \label{fig:base-point_all_dataset_grad_models}
    \end{subfigure}
    
    \caption{Kernel density estimate plot of the fraction of the explanation in the tangent space with \textbf{(a)} different base-point choices and \textbf{(b)} different gradient explainability models. The fraction of explanation in the tangent space is measured with $\mu_{x}$ (Equation \ref{def:mu}). The vertical line represents the expected fraction a random vector lies in the tangent space $\approx \sqrt{n/d}$, where $n =\dim(T_{x}M)$ and $d = \dim(M)$. On CIFAR10 and FER2013, $n = 144$. On MNIST32 and Fashion-MNIST, $n = 10$.}
    \label{fig:three_graphs}
\end{figure}


\section{Numerical Analysis}

In this section we approximate tangential base-point choices on four well-known datasets in computer vision: MNIST \cite{mnist}, Fashion-MNIST \cite{fashion}, CIFAR10 and FER2013 \cite{fer2013}. We demonstrate that the four common base-point choices defined in Section \ref{sec:common_choices} consistently provide explanations that are not well aligned with the tangent space. We further demonstrate tangentially aligned IG provides higher tangentially aligned explanations than three gradient explainability models: Gradient \cite{simonyan2014deepinsideconvolutionalnetworks}, Smooth Grad (SG) \cite{smilkov2017smoothgrad} and Input*Gradient (I*G) \cite{shrikumar2017justblackboxlearning}.

\subsection{Approximating the Tangent and Normal Space} Following \cite{brodt} the tangent space is approximated via a convolutional autoencoder. As discussed in \cite{riemannian_geom} if we consider the decoder, $\mathrm{dec}:L \to M$,
as a map from the latent space $L$ to the manifold $M$, then the Jacobian of the decoder is a linear map from the tangent spaces of $L$ and $M$
\begin{equation}
    J_{\mathrm{dec}}(x) : T_{x}L \to T_{\mathrm{dec}(x)}M.
\end{equation}
The Jacobian of the decoder can be computed via back-propagation \cite{riemannian_geom}. The tangent space of $M$ is spanned by the gradient of $\mathrm{dec}$ \cite{brodt}. For our work we require the normal space $T_{x}M^{\perp}$. Given a basis for the tangent space $\{\tau_{1},\ldots,\tau_{n}\}$, one can compute a basis for the normal space by 
\begin{equation}
\mathrm{Null}\left(\tau_{1},\ldots,\tau_{n}\right),
\end{equation}
where one considers the basis of the tangent space as a matrix.

\subsection{Experimental Setup}

We utilise the implementation of \cite{brodt} to generate the tangent space with a convolutional autoencoder and train a CNN for classification. The convolutional autoencoder has two convolutional layers with pooling followed by a fully connected layer with ReLU activation. A two layer CNN of kernel size 3 with dropout and Relu activation is used to perform image classification. Using the parameters of \cite{brodt}, $n = \dim(T_{x}M) = 144$ for CIFAR10 and FER2013 and for MNIST32 and Fashion $n = 10$. 
Explainability models are produced with the PyTorch library Captum.ai \cite{kokhlikyan2020captumunifiedgenericmodel}. 
\subsection{Complexity Analysis}
The problem of finding a base-point that gives tangentially aligned explanations can be phrased as

\begin{equation}
\label{eqn:optm}
    \alpha^*_{x} = \argmin_{x \neq \alpha} E_{x}(\alpha).
\end{equation}
If we suppose $M \subseteq \mathbb{R}^{d}$ is compact, then by Weierstrass's theorem such an $\alpha^{*}$ exists \cite{rudin}. The continuity of $E_{x}$ follows from the continuity of IG and norms. The condition $x \neq \alpha$ is required to ensure non-trivial solutions. A solution to the optimisation problem in Equation \ref{eqn:optm} can be approximated via gradient-descent. We note that zero, Gaussian and Uniform base-points have constant time $\mathcal{O}(1)$ complexity. Maximum $\ell_{2}$ distance is $\mathcal{O}(|D|)$ where $|D|$ is the number of points in the dataset. Calculating a tangential base-point has complexity $\mathcal{O}(\varepsilon)$, where $\varepsilon$ is the number of iterations in gradient descent to solve Equation \ref{eqn:optm}. IG with base-point $\alpha_{x}^{*}$ from Equation \ref{eqn:optm} will be referred to as tangentially aligned IG.
 \begin{figure*}[!htb]
  \centering
  \includegraphics[width = \linewidth]{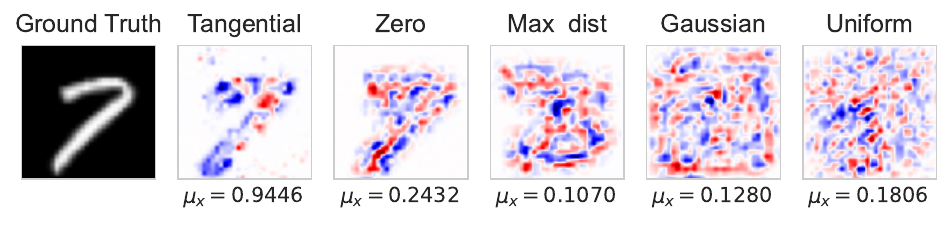}
    \includegraphics[width = \linewidth]{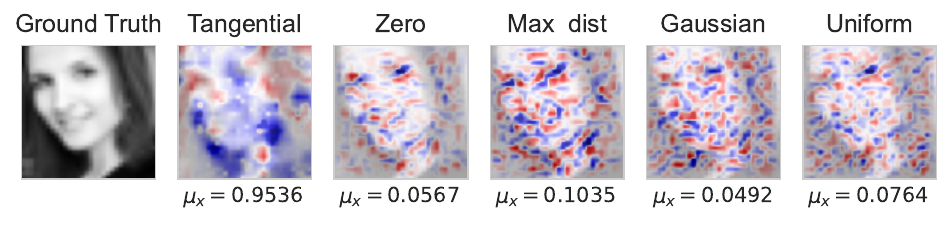}
      \includegraphics[width =\linewidth]{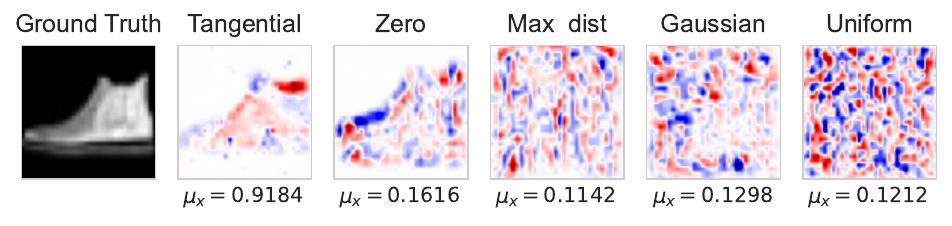}

  \caption{Attributions of IG with differing base-point choice on example points from MNIST, FER2013, and Fashion-MNIST. The fraction of explanation in the tangent space is denoted by $\mu_{x}$ (Equation \ref{def:mu}).}
  \label{fig:base-point_mnist}
\end{figure*}
\subsection{Comparison of Different Base-points with Tangential Integrated Gradients}
For each dataset IG is applied to the CNN with the base-points defined in Section \ref{sec:common_choices} and the fraction of each explanation is calculated via Equation \ref{def:mu}. To calculate each base-point in Section \ref{sec:common_choices} we use the implementation provided by \cite{sturmfels2020visualizing}.

For each point we approximate the solution to Equation \ref{eqn:optm} to provide tangential explanations on each dataset. To approximate the solution to Equation \ref{eqn:optm} over all points we use the same learning rate and number of iterations. Some points may require a different learning rate and number of iterations to achieve higher tangential alignment. We leave this to future work. In Figure \ref{fig:base-point_all_dataset} we have the distributions of the fraction in the tangent space on FER2013, CIFAR10, MNIST32 and Fashion MNIST. We see in Figure \ref{fig:base-point_all_dataset}  that approximating solutions to Equation \ref{eqn:optm} consistently provides explanations with high tangential alignment.

We see in Figure \ref{fig:base-point_all_dataset} that the uniform base-point provides explanations consistently close to the normal space; followed by maximum $\ell_2$ distance and Gaussian. We note that on FER2013 and CIFAR10 the Gaussian base-point performs better than zero, uniform, and maximum $\ell_2$ distance. The better performance of a Gaussian base-point is likely due to the smoothing parameter $\sigma$ defined in Section \ref{sec:common_choices}. It is the goal of future work to determine the impact of $\sigma$ on tangential alignment. The vertical lines in Figure \ref{fig:base-point_all_dataset} indicate the expectation a random vector will lie in the tangent space. The expectation is approximately $\sqrt{n/d}$, where $n$ and $d$ are the dimensions of the tangent space approximation and manifold, respectively \cite{brodt}. 
An explanation is therefore sufficiently aligned with the tangent space when that fraction in the tangent space is greater than $\sqrt{n/d}$. We see in Figure \ref{fig:base-point_all_dataset} that standard base-point choices on CIFAR10 are significantly below the vertical line. It is the goal of future work to determine if the dimension of the tangent space of CIFAR10 of $n = 144$ or the parameter of the Gaussian base-point impacts the tangential alignment of IG on CIFAR10.

We provide in Figure \ref{fig:base-point_mnist}, example integrated gradient explanations for a point on MNIST32, FER2013, and Fashion-MNIST with differing base-point choice. We see that our method provides tangentially aligned explanations with $\mu_{x} > 0.91$ for all datasets. The tangentially aligned integrated gradient attributions are clear and perceptually aligned with the object to classify in the image. We see in Figure \ref{fig:base-point_mnist} that uniform, maximum $\ell_2$ distance, and Gaussian are consistently random noise.


\subsection{Comparison of Gradient Explainability Models with Tangential Integrated Gradients}

In this section we compare tangentially aligned integrated gradients with three common gradient explainability models: Gradient, Smooth Grad and Input *Gradient. The aforementioned gradient explainability models do not require a base-point choice. We demonstrate that tangentially aligned integrated gradients significantly improves upon integrated gradients. The gradient explainability models for a given model are defined as follows:

\begin{enumerate}
    \item \textbf{Gradient} The gradient of a model $f$ at $x \in \R^{d}$ for class $i$ is defined as:
    \begin{equation}
        \mathrm{grad(x)_{i} \coloneqq \frac{\partial f(x)_i}{\partial x}}.
    \end{equation}
    \item \textbf{Smooth Grad}
    We define Smooth Grad with $n$ samples and standard deviation $\sigma$ as:
    \begin{equation}
        \mathrm{SmoothGrad}(x) = \frac{1}{n}\sum_{i = 1}^{n} \nabla f(x + a),
    \end{equation}
    where, $a \sim \mathcal{N}(0,\sigma^2)$. Following \cite{brodt} we take $\sigma = 0.02$ and $n = 25$.
    \item \textbf{Input*Gradient} Input*Gradient is defined as:
    \begin{equation}
        \mathrm{Input*Gradient} \coloneqq x\odot \frac{\partial f(x)_i}{\partial x}.
    \end{equation}
\end{enumerate}

In Figure \ref{fig:base-point_all_dataset_grad_models} we provide density plots of the fraction an attribution is in the tangent space for: Gradient, Smooth Grad, Input*Gradient and tangentially aligned integrated gradients. We see in Figure \ref{fig:base-point_all_dataset_grad_models} that tangentially aligned integrated gradients provides attributions consistently in the tangent space, out-performing the aforementioned gradient explainability models. In Figures \ref{fig:base-point_all_dataset} and \ref{fig:base-point_all_dataset_grad_models}  Gradient, Smooth Grad and Input*Gradient provide better tangential alignment than the common base-point choices provided in Section \ref{sec:common_choices} on MNIST and CIFAR10. On Fashion-MNIST we see that the zero base-point choice provides comparable performance with Gradient, Smooth Grad and Input*Gradient. In Figures \ref{fig:base-point_all_dataset} and \ref{fig:base-point_all_dataset_grad_models}  we see that on FER2013, Gradient, Smooth Grad and Input*Gradient perform similarly to Gaussian, maximum $\ell_{2}$ distance and zero base-point choices for integrated gradients. All gradient models on FER2013 outperform the uniform base-point choice for integrated gradients. We see in Figures \ref{fig:base-point_all_dataset} and  \ref{fig:base-point_all_dataset_grad_models}, Gradient, Smooth Grad, and Input*Gradient tend to out-perform Integrated gradients with standard the standard base-point choices. Tangential integrated gradients out-performs the aforementioned gradient explainability models and standard base-point choices.

\section{Conclusions and Future Work}
In this work we investigated how to choose base-points for IG that provide tangentially aligned explanations. We provided theoretical conditions for a base-point to provide tangentially aligned explanations for any BAM. We demonstrated how to numerically approximate the base-point which provides tangentially aligned explanations and validated this approach on several well-known image classification datasets. In future work we seek to further investigate the theoretical conditions a base-point must have to provide tangential explanations.

\begin{acknowledgments}
The Commonwealth of Australia (represented by the Defence Science and Technology Group) supports this research through a Defence Science Partnerships agreement.
Lachlan Simpson is supported by a scholarship from the University of Adelaide. 
\end{acknowledgments}

\bibliography{sample-ceur}

\end{document}